\documentclass[10pt,twocolumn,letterpaper]{article}

\usepackage{cvpr}
\usepackage{times}
\usepackage{epsfig}
\usepackage{graphicx}
\usepackage{amsmath}
\usepackage{amssymb}

\usepackage{microtype}
\usepackage{multirow}
\usepackage{enumitem}
\usepackage{eso-pic}
\usepackage[linesnumbered,ruled]{algorithm2e}
\usepackage{textcomp}
\usepackage{epsfig}
\usepackage{subcaption}
\usepackage{booktabs}
\usepackage{enumitem}

\usepackage[breaklinks=true,bookmarks=false]{hyperref}

\cvprfinalcopy 

\def\cvprPaperID{****} 

\begin{document}

\title{Jointly Optimize Data Augmentation and Network Training:\\Adversarial Data Augmentation in Human Pose Estimation}


\author{
Xi Peng\thanks{Contributed equally. The project page is publicly available: \url{https://sites.google.com/site/xipengcshomepage/cvpr2018}}\\
Rutgers University\\
{\tt\small xipeng.cs@rutgers.edu}
\and
Zhiqiang Tang\footnotemark[1]\\
Rutgers University\\
{\tt\small zt53@cs.rutgers.edu}
\and
Fei Yang\\
Facebook\\
{\tt\small yangfei@fb.com}
\and
Rogerio S. Feris\\
IBM T.J. Watson Research Center\\
{\tt\small rsferis@us.ibm.com}
\and
Dimitris Metaxas\\
Rutgers University\\
{\tt\small dnm@cs.rutgers.edu}
}

\maketitle

\begin{abstract}

Random data augmentation is a critical technique to avoid overfitting in training deep neural network models. However, data augmentation and network training are usually treated as two isolated processes, limiting the effectiveness of network training. Why not jointly optimize the two? We propose adversarial data augmentation to address this limitation. The main idea is to design an augmentation network (generator) that competes against a target network (discriminator) by generating ``hard'' augmentation operations online. The augmentation network explores the weaknesses of the target network, while the latter learns from ``hard'' augmentations to achieve better performance. We also design a reward/penalty strategy for effective joint training. We demonstrate our approach on the problem of human pose estimation and carry out a comprehensive experimental analysis, showing that our method can significantly improve state-of-the-art models without additional data efforts.


\end{abstract}

\section{Introduction}

Deep Neural Networks (DNNs) have achieved significant improvements in many computer vision tasks \cite{alex2012alexnet, girshick2014rich, jaderberg2015spatial, Mohamed17}. A key ingredient for the success of state-of-the-art deep learning models is the availability of large amounts of training data. However, data collection and annotation are costly, and for many tasks, only a few training examples may be available. In addition, natural images usually follow a long-tail distribution \cite{zhu2014capturing,tang2015face}. Effective training examples that lead to more robust classifiers may still be rare even if a large amount of data have been collected.

A common solution for this problem is to perform random data augmentation \cite{lecun1998gradient, toshev2014deeppose}. Prior to being fed into the network, training images are heuristically jittered by predefined transformations ({\it e.g.}, scaling, rotating, occluding) to increase variations. This strategy is simple, but data augmentation and network training are still treated as isolated processes, leading to the following issues.

\begin{figure}[t]
\centering
\includegraphics[width=.48\textwidth]{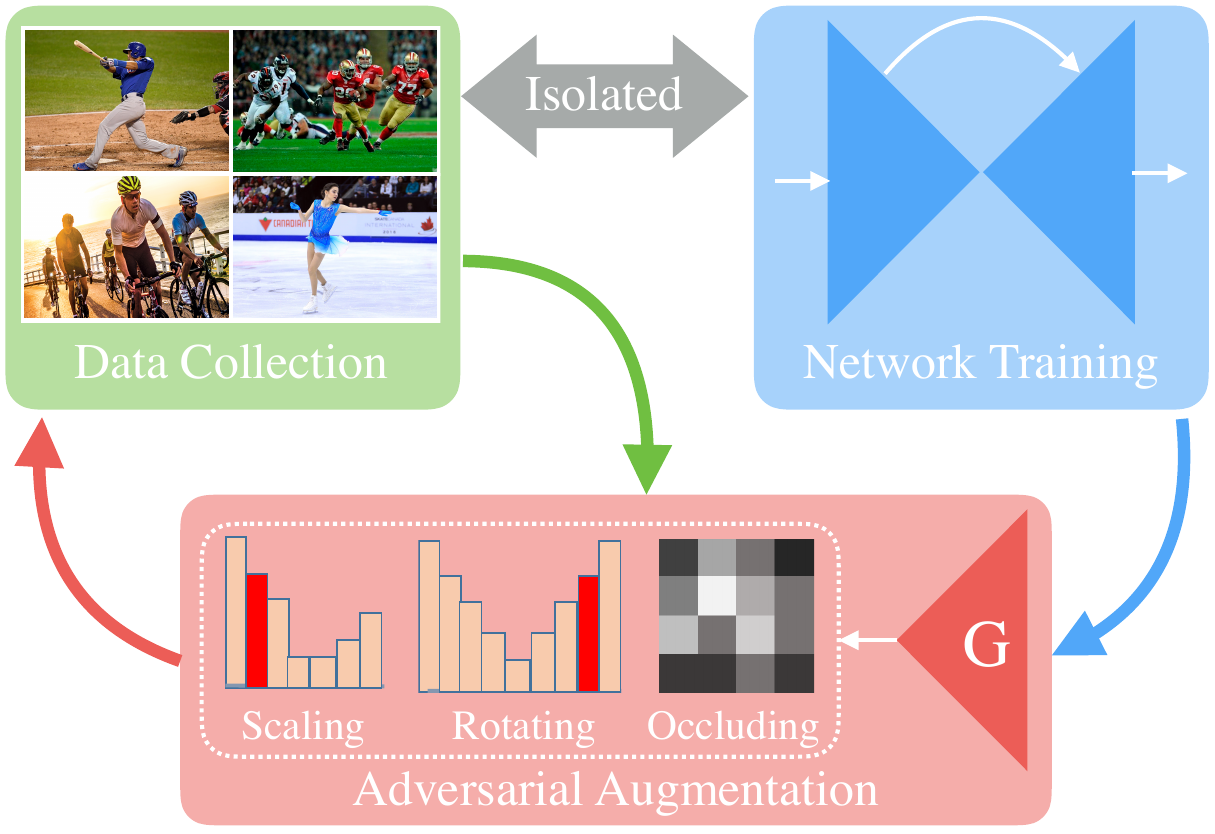}
\caption{Data preparation and network training are usually isolated. We propose to bridge the two by generating adversarial augmentations online. The generations are conditioned to both training images and the status of the target network.}
\label{fig:intro}
\end{figure}

First, the entire training set is usually applied the same random data augmentation strategy without considering the individual difference. This may produce many ineffective variations that are either too ``hard'' or too ``easy'' to help the network training \cite{shrivastava2016training,wang2017fast}. Second, random data augmentations can hardly match the dynamic training status since they are usually sampled from static distributions. Third, Gaussian distribution are widely used, which cannot address the long-tail issue since there would be a small chance to sample rare but useful augmentations.

A natural question then arises: can data augmentation and network training be jointly optimized, so that effective augmentations can be generated online to improve the training?  

In this work, we answer the above question by proposing a new approach that leverages adversarial learning for joint optimization of data augmentation and network training (see Figure \ref{fig:intro}). Specifically, we investigate the problem of human pose estimation, aiming to improve the network training with bounded datasets. Note that our approach can be generalized to other vision tasks, such as face alignment \cite{peng2016recurrent} and instance segmentation \cite{long2015fully, he2017maskrcnn}.

Given an off-the-shelf pose estimation network, our goal is to obtain improved training from a bounded dataset. Specifically, we propose an augmentation network that acts as a generator. It aims to create ``hard'' augmentations that intend to make the pose network fail. The pose network, on the other hand, is modeled as a discriminator. It evaluates the quality of the generations, and more importantly, tries to learn from the ``hard'' augmentations. The main idea is to generate adversarial data augmentations online, conditioned to both input images and the training status of the pose network. In other words, the augmentation network explores the weaknesses of the pose network which, at the same time, learns from adversarial augmentations for better performance.

\begin{figure*}[t!]
\minipage{0.7\textwidth}
    \centering
  \includegraphics[width=0.95\linewidth]{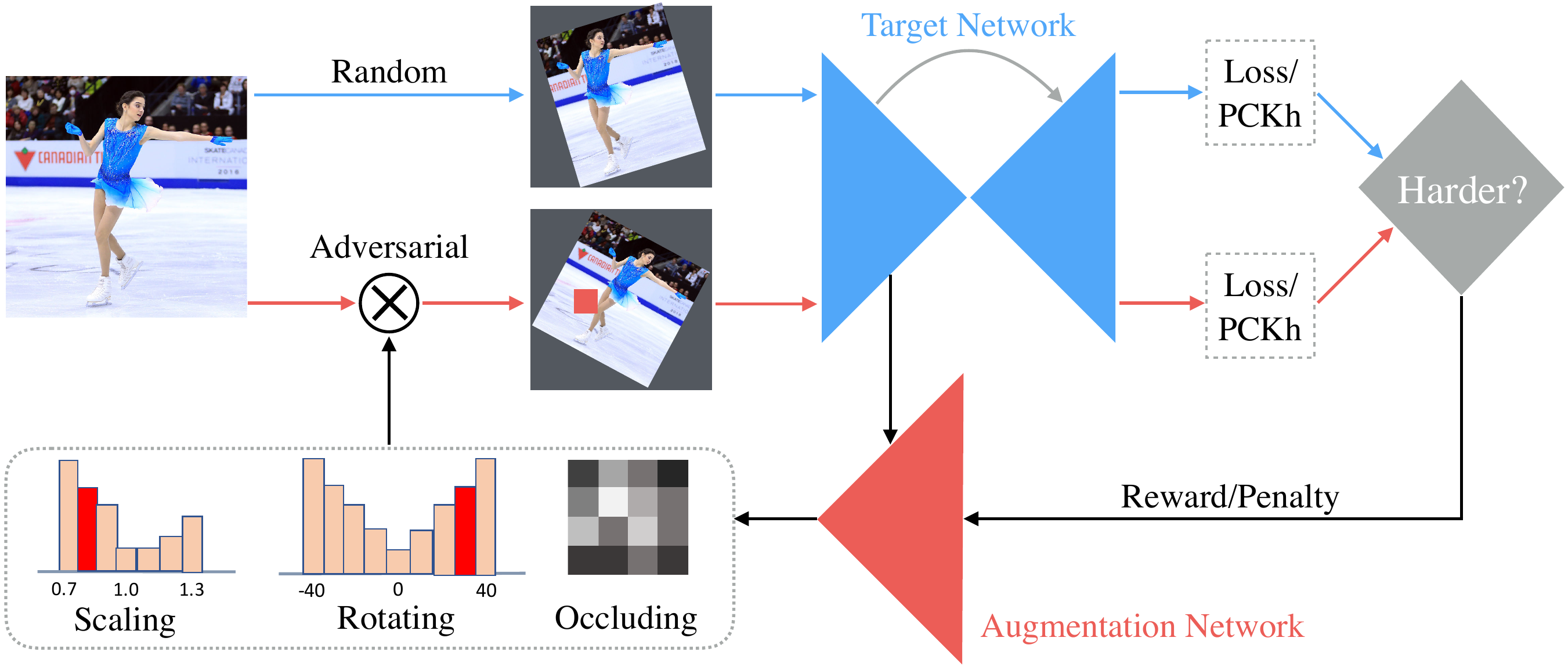}
\endminipage
\minipage{0.29\textwidth}
    \centering
  \includegraphics[width=0.99\linewidth]{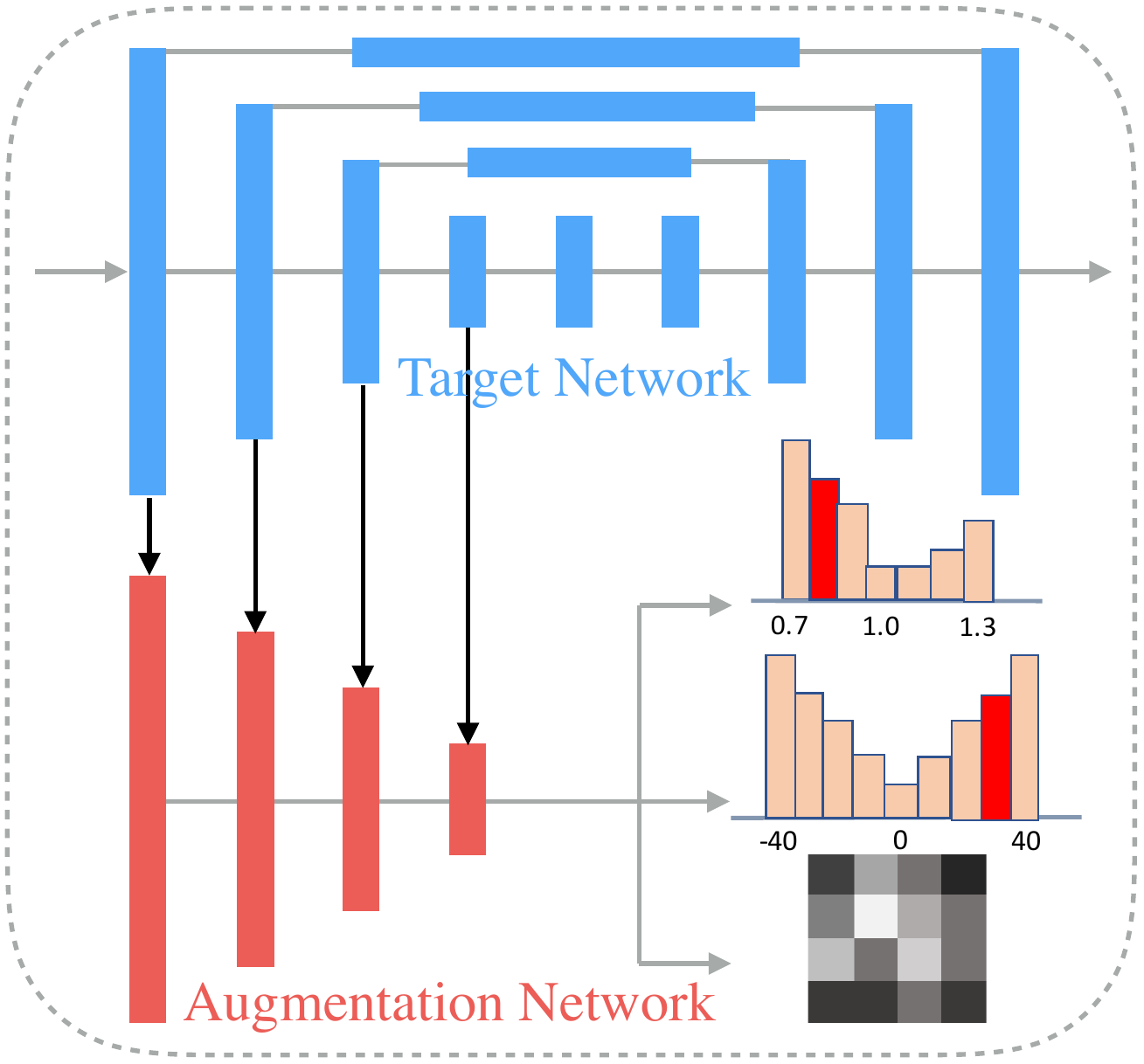}
\endminipage
\caption{{\bf Left}: Overview of our approach. We propose an augmentation network to help the training of the pose network. The former creates hard augmentations; the latter learn from generations and produces reward/penalty for model update. {\bf Right}: Illustration of the augmentation network. Instead of raw images, it takes hierarchical features of an U-net as inputs.}
\label{fig:overview}
\end{figure*}

Jointly optimizing the two networks is a non-trivial task. Our experiments indicate that a straightforward design, such as directly generating adversarial pixels \cite{goodfellow2014gan, reed2016generative} or deformations \cite{wang2017fast, jaderberg2015spatial}, would yield problematic convergence behaviors ({\it e.g.} divergence and model collapse). Instead, the augmentation network is designed to generate adversarial distributions, from which augmentation operations ({\it i.e.} scaling, rotating, occluding) are sampled to create new data points. Besides, we propose a novel reward and penalty policy to address the issue of missing supervisions during the joint training. Moreover, instead of a raw image, the augmentation network is designed to take the byproduct, {\it i.e.} hierarchical features, of the pose network as the input. This can further improve the joint training efficiency using additional spatial constraints. To summarize, our key contributions are:

\begin{itemize}
    \item To the best of our knowledge, we are the first to investigate the joint optimization of data augmentation and network training in human pose estimation.
    
    \item We propose an augmentation network to play a minimax game against the target network, by generating adversarial augmentations online.
    
    \item We take advantage of the wildly used U-net design and propose a reward and penalty policy for the efficient joint training of the two networks.
    
    \item Strong performance on public benchmarks, {e.g.} MPII and LSP, as well as intensive ablation studies, validate our method substantially in various aspects.
\end{itemize}

\section{Related Work}

We provide a brief overview of previous methods that are most relevant to ours in three categories.

{\bf Adversarial learning.} Generative Adversarial Networks (GANs) \cite{goodfellow2014gan,zhang2017image,yizhe_zsl_2018} are designed as playing minimax games between generator and discriminator. Yu and Grauman \cite{yu2017semantic} use GANs to synthesize image pairs to overcome the sparsity of supervision when learning to compare
images. A-Fast-RCNN \cite{wang2017fast} uses GANs to generate deformations for object detection. Recent applications of GANs in human pose estimation include \cite{yu2017adversarial} and \cite{chou2017self}. They both treat the pose estimation network as the generator and use a discriminator to provide additional supervision. However, in our design, the pose estimation network is treated as a discriminator, while the augmentation network is designed as a generator to create adversarial augmentations.

{\bf Hard example mining.} It is wildly used in training SVM models for object detection \cite{uijlings2013selective,wu2016towards,shrivastava2016training}. The idea is to perform an alternative optimization between model training and data selection. Hard example mining focuses on how to select hard examples from the training set for effective training. It cannot create new data that do not exist in the training set. In contrast, we propose an augmentation network (generator) to actively generate adversarial data augmentations. This will create new data points that may not exist in the training set to improve the pose network (discriminator) training. 

{\bf Human pose estimation.} 
DeepPose \cite{toshev2014deeppose} proposed to use deep neural networks for human pose estimation. Since then, deep learning based methods started to dominate this area \cite{carreira2016human, tompson2015efficient, hu2016bottom, pishchulin2016deepcut, lifshitz2016human, gkioxari2016chained, insafutdinov2016deepercut, wei2016convolutional, bulat2016human, newell2016stacked}.
For instance, Tompson {\it et al.} \cite{tompson2014joint} used multiple branches of convolutional networks to fuse the features from an image pyramid. They applied Markov Random Field for post-processing. Chen {\it et al.} \cite{chen2014articulated} also tried to combine neural networks with the graphical model inference to improve the pose estimation accuracy. 

Recently, cascade models become popular for human pose estimation. They usually connect a series of deep neural networks in cascade to improve the estimation in a stage-by-stage manner. For example, {\em Convolutional Pose Machines} \cite{wei2016convolutional} brings obvious improvements by cascading multiple networks and adding intermediate supervisions. Better performance is achieved by the stacked hourglass network architecture \cite{newell2016stacked}, which also relies on multi-stage pose estimation. More recently, Chu {\it et al.} \cite{chu2017multi} added some layers into the stacked hourglass network for attention modeling. Yang {\it et al.} \cite{yang2017learning} also enhanced its performance by using pyramid residual modules. In this paper, instead of designing a new pose estimation network, we are more interested in how to jointly optimize data augmentation and network training. So we can obtain improved training effect on any off-the-shelf deep neural network without looking for more data.

\section{Adversarial Data Augmentation} \label{sec:ada}
Given a pre-designed pose network, {\it e.g.} the stacked hourglass pose estimator \cite{newell2016stacked}, our goal is to improve its training without looking for more data. Random data augmentation is widely used in deep neural network training. However, random data augmentations that are sampled from static distributions can hardly follow the dynamic training status, which may produce many ineffective variations that are either too ``hard'' or too ``easy'' to help the network training \cite{shrivastava2016training,wang2017fast}.

Instead, we propose to leverage adversarial learning to optimize the data augmentation and the network training jointly. The main idea is to learn an augmentation network $G(\cdot|\theta_G)$ that generates ``hard'' augmentations that may increase the pose network loss. The pose network $D(\cdot|\theta_D)$, on the other hand, tries to learn from the adversarial augmentations and, at the same time, evaluates the quality of the generations. Please refer to Figure \ref{fig:overview} for an overview of our approach.

{\bf Generation path.} The augmentation network is designed as a generator. It outputs a set of distributions of augmentation operations. Mathematically, the augmentation network $G$ outputs adversarial augmentation $\tau_{a}(\cdot)$ that may increase $D$'s loss, compared with random augmentation $\tau_{r}(\cdot)$, by maximizing the expectation:
\begin{equation}
\label{eq:G}
\max_{\theta_{G}} \; \underset{\mathbf{x}\sim\Omega}{\mathbb{E}}
\underset{ \substack{\tau_{r}\sim{\Gamma} \\  \tau_{a}\sim{G(\mathbf{x},\theta_D)}} }{\mathbb{E}}
\mathcal{L}[D(\tau_{a}(\mathbf{x}), \mathbf{y})] - \mathcal{L}[D(\tau_{r}(\mathbf{x}), \mathbf{y})],
\end{equation}
where $\Omega$ is the training image set and $\Gamma$ is the random augmentation space. $\mathcal{L}(\cdot,\cdot)$ is a predefined loss function and $\mathbf{y}$ is the image annotation. We highlight $G(\mathbf{x},\theta_D)$ to specify that the generation of $G$ is conditioned to both the input image $\mathbf{x}$ and the current status of the target network $D$.

{\bf Discrimination path.} The pose network is designed as a discriminator. It plays two roles: 1) $D$ evaluates the generation quality as indicated in Equation \eqref{eq:G}; 2) $D$ tries to learn from adversarial generations for better performance by minimizing the expectation:
\begin{equation}
\label{eq:D}
\min_{\theta_{D}} \; \underset{\mathbf{x}\sim\Omega}{\mathbb{E}} \; \;
\underset{ \tau_{a}\sim{G(\mathbf{x},\theta_D)} }{\mathbb{E}} \;
\mathcal{L}[D(\tau_{a}(\mathbf{x}), \mathbf{y})],
\end{equation}
where adversarial augmentation $\tau_{a}$ can better reflect the weakness of $D$ than random augmentation $\tau_{r}$, resulting in more effective network training.

{\bf Joint training.} The joint training of $G$ and $D$ is a non-trivial task. Augmentation operations are usually not differentiable \cite{wang2017fast}, which stops gradients to flow from $D$ to $G$ in backpropagation. To solve this issue, we propose a reward and penalty policy to create online ground truth of $G$. So $G$ can always be updated to follow $D$'s training status. The details will be explained soon in Section \ref{sec:joint}.

It is crucial that $G$ generates distributions instead of direct operations \cite{wang2017fast} or adversarial pixels \cite{reed2016generative}. Our experiments indicate that, by sampling from distributions, the generation is more robust to outliers which may produce upside-down augmentations. Thus, there is less chance that $D$ would get trapped in a local optimum. 

{\bf Comparison with prior methods.} We want to stress that there is a sharp difference between our method and the recent adversarial human pose estimation techniques \cite{yu2017adversarial,chou2017self}. The latter usually follow a common design that connects a pose network (generator) with an additional network (discriminator) to obtain adversarial loss. In contrast, we propose to learn an adversarial network (generator) to improve the pose network (discriminator), by jointly optimizing data augmentation and network training.

Our method is also different from others that perform online hard example mining \cite{uijlings2013selective,shrivastava2016training}. Our method can create new data points that may not exist in the dataset, whereas the latter is usually bounded by the dataset. An exception is \cite{wang2017fast} that uses GANs to generate deformations for object detection. However, how to jointly optimize data augmentation and network training, especially for human pose estimation, is still an open question without investigation.

\section{Adversarial Human Pose Estimation}

Our task is to improve the training of a pre-designed pose network. We take the wildly used U-net design \cite{newell2016stacked,ronneberger2015unet} as an example. As illustrated in Figure \ref{fig:overview} (right), the augmentation network follows an encoder architecture. It takes the bridged features of the U-net as inputs instead of raw images for efficient training. A set of distributions are then generated to sample three typical augmentations: scaling, rotating, and hierarchical occluding. Furthermore, we propose a reward and penalty strategy for efficient joint training.

\subsection{Adversarial Scaling and Rotating (ASR)} \label{sec:asr}

\begin{figure}[t!]
\centering
\includegraphics[width=.47\textwidth]{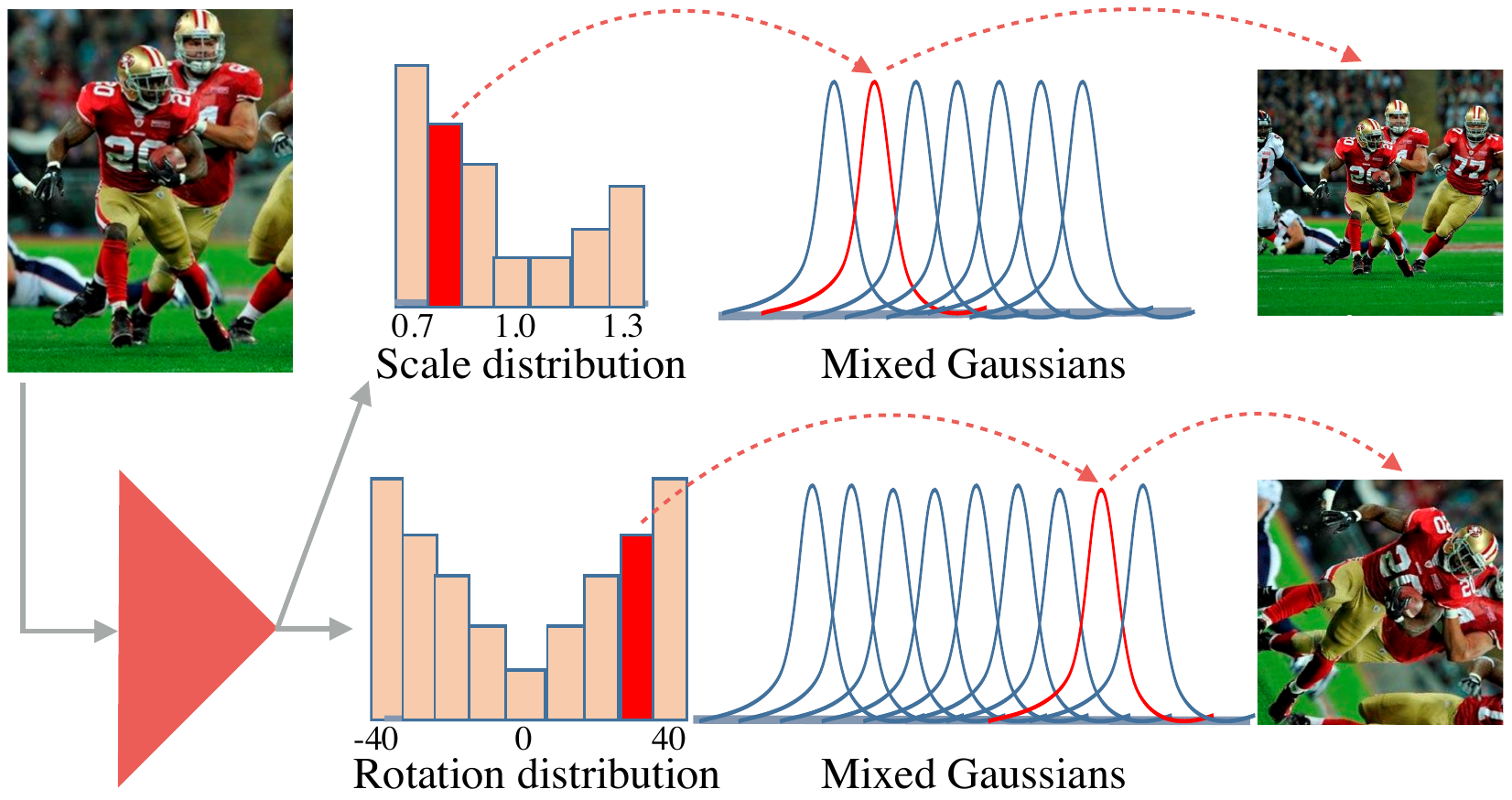}
\caption{Adversarial scaling and rotating. Our generator predicts distributions of mixed Gaussian, from which scaling and rotating are then sampled to augment the training image.}
\label{fig:asr}
\end{figure}

The augmentation network generates adversarial augmentations by scaling and rotating the training images. The pose network then learns from the adversarial augmentations for more effective training. In our experiments, we find that a direct generation would collapse the training. It would easily generate upside-down augmentations that are the hardest in most cases. Instead, we divide the augmentation ranges into $m$ and $n$ bins ({\it e.g.} $m=7$ for scaling and $n=9$ for rotating). Each bin corresponds to a small bounded Gaussian. The augmentation network will first predict distributions over scaling and rotating bins. Then, the corresponding Gaussian is activated by sampling from distributions. Please refer to Figure \ref{fig:asr} for an illustration of the sampling process.

{\bf ASR pre-training.} It is crucial to pre-train the augmentation network so it can obtain the sense of augmentation distributions before the joint training. For every training image, we can sample totally $m \times n$ augmentations, each of which is drawn from a pair of Gaussians. The augmentations are then fed forward into the target network to calculate the loss which represents how ``difficult'' the augmentation is. We accumulate $m \times n$ losses into the corresponding scaling and rotation bins. By normalizing the sum of bins to 1, we generate two vectors of probabilities: $P^s \in \mathbb{R}^m$ and $P^r \in \mathbb{R}^n$, which approximate the ground truth of scaling and rotation distributions, respectively.

Given the ground-truth distributions $P^s$ and $P^r$, we propose a KL-divergence loss to pre-train the augmentation network for scaling and rotating:
\begin{equation}\label{eq:loss-ASR}
    \mathcal{L}_{SR}=\sum_{i=1}^mP_i^s \log\frac{P_i^s}{\tilde{P}_i^s}+\sum_{i=1}^nP_i^r \log\frac{P_i^r}{\tilde{P}_i^r},
\end{equation}
where $\tilde{P}^s \in \mathbb{R}^m$ and $\tilde{P}^r \in \mathbb{R}^n$ are the predicted distributions following the above generation procedure. $m$ and $n$ are the numbers of scale and rotation bins.

{\bf Discussion.} Predicting distributions instead of direct augmentations has two advantages. First, it introduces uncertainties to avoid upside-down augmentations during the pre-training. Second, it helps to address the issue of missing ground truth during the joint training, which will be explained in Section \ref{sec:joint}. In our design, the scaling and rotating are directly applied on training images instead of deep features \cite{wang2017fast}. The reason is we want to preserve the location correspondence between image pixels and landmark coordinates. Otherwise, we might hurt the localization accuracy once the intermediate feature maps are disturbed.

\subsection{Adversarial Hierarchical Occluding (AHO)} \label{sec:aho}

In addition to scaling and rotating, the augmentation network also generates occluding operations to make the task even ``harder''. The human body has a linked structure where joint locations are highly correlated to each other. By occluding parts of the image, the pose network is encouraged to learn strong references among visible and invisible joints \cite{peng2015circle}.

\begin{figure}[t!]
\centering
\includegraphics[width=.44\textwidth]{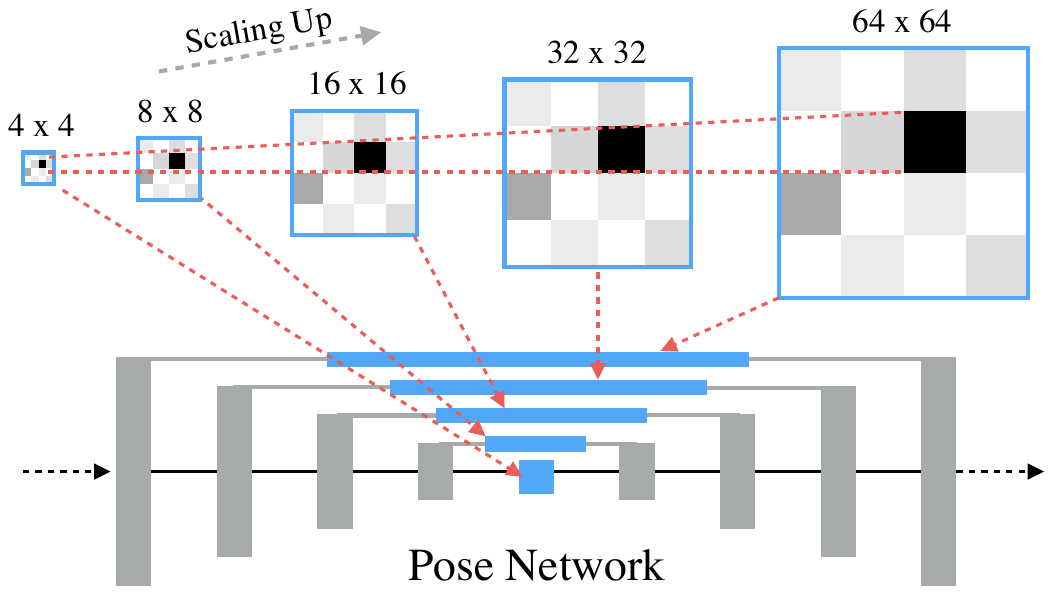}
\caption{Adversarial Hierarchical Occluding. The occlusion mask is generated at the lowest resolution and then scaled up to apply on hierarchical bridge features of the pose network.}
\label{fig:aho}
\end{figure}

Different from scaling and rotating, we find that it is more effective to occlude deep features instead of image pixels. It does not have the location correspondence issue since joint positions are unchanged after the occluding. Specifically, the augmentation network generates a mask indicating which part of features to be occluded so that the pose network has more estimation errors. We only generate the mask at the lowest resolution of $4 \times 4$. The mask is then scaled up to $64 \times 64$ to apply on bridge features of the U-net. Figure \ref{fig:aho} explains the proposed hierarchical occluding.

{\bf AHO pre-training.} Similar to scaling and rotating, the augmentation network predicts an occluding distribution instead of an instance occluding mask. The first task is to create the ground truth of the occluding distribution. The idea is to assign values into a grid of $w \times h$ ({\it e.g.} $w=h=4$). The value indicates the importance of the features at the corresponding cell. To achieve this, we vote a joint to one of the $w \times h$ cells according to its coordinates. By counting all joints from all images and normalizing the sum of cells to 1, we generate a heat map $P^o \in \mathbb{R}^{w \times h}$, which approximates the ground truth of the occluding distribution.

Given the ground-truth distribution $P^o$, we propose a KL-divergence loss to pre-train the AHO task:
\begin{equation} \label{eq:loss-AHO}
\mathcal{L}_{AHO}=\sum_{i=1}^h\sum_{j=1}^wP_{i,j}^o \log\frac{P_{i,j}^o}{\tilde{P}_{i,j}^o},
\end{equation}
where $\tilde{P}^o \in \mathbb{R}^{w \times h}$ is the heat map predicted by the augmentation network. To generate the occluding mask, we sample one or two cells according to $\tilde{P}^o$, which are labeled as 0 while the rests are labeled as 1. 

{\bf Discussion.} Intuitively, there are three ways to apply hierarchical occluding: (1) a single mask scales up from the lowest to the highest resolutions, (2) a single mask scales down from the highest to the lowest resolutions, and (3) independent masks are generated at different resolutions. We exclusively use the first design in our approach since it would occlude more than needed due to the large receptive field in the second case, and the occluded information may be compensated at other resolutions in the third case.

\begin{algorithm}[t!]
\SetAlgoLined
\KwIn{Mini-batch $\mathbf{X}$, augmentation net $G$, pose net $D$.}
\KwOut{G, D.}
Randomly and equally divide $\mathbf{X}$ into $\mathbf{X}_1$, $\mathbf{X}_2$ and $\mathbf{X}_3$;

Train D using $\mathbf{X}_1$;

Train D, G using $\mathbf{X}_2$ with ASR following Alg. \ref{alg:learning};

Train D, G using $\mathbf{X}_3$ with AHO following Alg. \ref{alg:learning};
\caption{Training scheme of a mini batch} \label{alg:jointtrain}
\end{algorithm}

\subsection{Joint Training of Two Networks} \label{sec:joint}

Once ASR and AHO are pre-trained, we can jointly optimize the augmentation network and the pose network. As we mentioned in Sec. \ref{sec:ada}, this is a non-trivial task since the augmentation ground truth is missing. A naive approach could be repeating the pre-training process as described in Section \ref{sec:asr} and Section \ref{sec:aho} online. However, it would be extremely time-consuming since there are a large number of augmentation combinations.

{\bf Reward and penalty.} Instead, we propose a reward and penalty policy to address this issue. The key idea is, the prediction of the augmentation network should be updated according to the current status of the target network, while its quality should be evaluated by comparing with a reference. 

To this end, we sample a pair of augmentations for each image: 1) an adversarial augmentation $\tau_a$ and 2) a random augmentation $\tau_r$, as indicated in Equation \eqref{eq:G}. If the adversarial augmentation is harder than the random one, we reward the augmentation network by increasing the probability of the sampled bin (ASR) or cell (AHO). Otherwise, we penalize it by decreasing the probability accordingly.

Mathematically, let $\tilde{P} \in \mathbb{R}^k$ denotes the predicted distribution of the augmentation network. $P \in \mathbb{R}^k$ denotes the ground truth we are looking for. $k$ is the number of bins (ASR) or cells (AHO) and $i$ is the sampled one.

If the adversarial augmentation $\tau_a$ leads to higher pose network loss (more ``difficult'') comparing with the reference (a random augmentation $\tau_r$), we update $P$ by rewarding:
\begin{equation} \label{eq:reward}
P_{i} = \tilde{P}_{i} + \alpha \tilde{P}_i; \quad P_{j} = \tilde{P}_{j} - \frac{\alpha \tilde{P}_i}{k-1}, \forall j \neq i.
\end{equation}
Similarly, if $\tau_a$ leads to lower pose network loss (less ``difficult'') comparing with $\tau_r$, we update $P$ by penalizing:
\begin{equation} \label{eq:penalty}
P_{i} = \tilde{P}_{i} - \beta \tilde{P}_i; \quad P_{j} = \tilde{P}_{j} - \frac{\beta \tilde{P}_i}{k-1}, \forall j \neq i,
\end{equation}
where $0 < \alpha, \beta \leq 1$ are hyperparameters that controls the amount of reward and penalty. The augmentation network keeps updating online, regardless of being rewarded or penalized, generating adversarial augmentations that intend to improve the pose network.

\begin{algorithm}[t!]
\SetAlgoLined
\KwIn{Image $\mathbf{x}$, augmentation network $G$, pose network $D$.}
\KwOut{G, D.}
Forward $D$ to get bridge features $\mathbf{f}$;

Forward $G$ with $\mathbf{f}$ to get a distribution $P$;

Sample an adversarial augmentation $\tilde{\mathbf{x}}$ from $P$;

Forward $D$ with $\tilde{\mathbf{x}}$ to compute loss $\tilde{\mathcal{L}}$;

Random augment $\mathbf{x}$ to get $\hat{\mathbf{x}}$;

Forward $D$ with $\hat{\mathbf{x}}$ to compute loss $\hat{\mathcal{L}}$;

Compare $\tilde{\mathcal{L}}$ with $\hat{\mathcal{L}}$ to update $G$ using \eqref{eq:loss-ASR} and \eqref{eq:loss-AHO};

Update $D$;
\caption{Training scheme of one image.}  \label{alg:learning}
\end{algorithm}

{\bf Discussion.} The pose network can learn from the ordinary random augmentation to maintain its regular performance. More importantly, it can also learn from the adversarial augmentations to achieve better performance. The adversary augmentations may become too hard for the pose network if we apply ASR and AHO simultaneously. Thus, we alternately apply ASR and AHO on different images. Here we equally split every mini batch into three shares: one performs the random data augmentation, one performs ASR augmentation, and one performs AHO augmentation. Please check Algorithm \ref{alg:jointtrain} for the details.

\begin{figure*}[t!]
\centering
\includegraphics[width=0.95\textwidth]{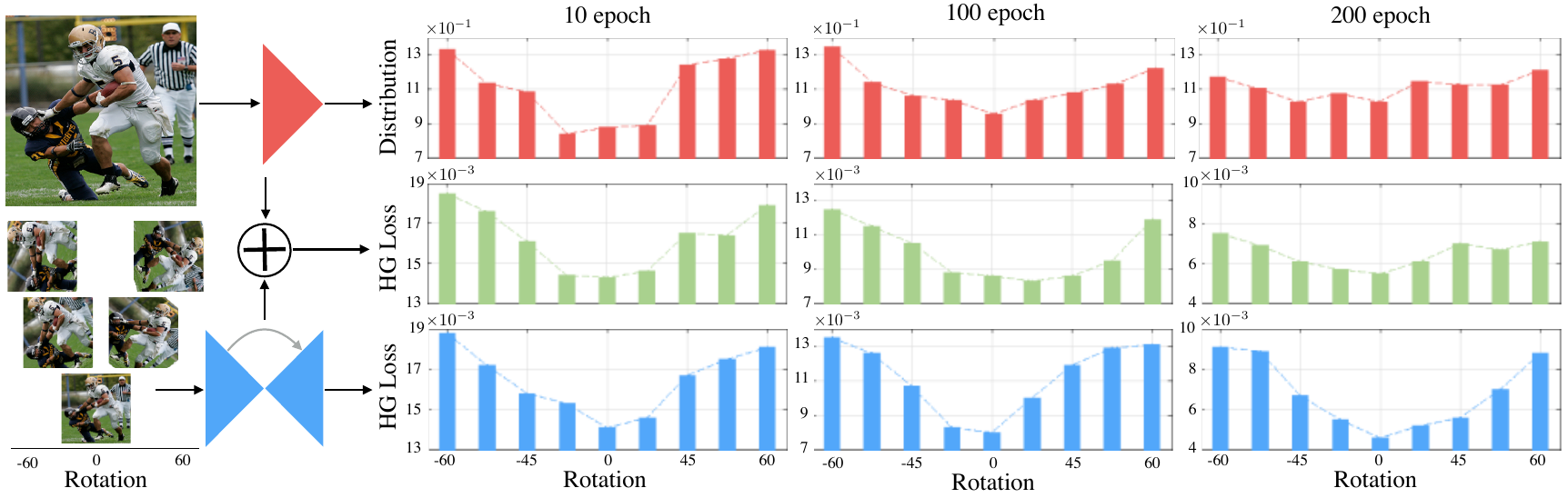}
\caption{Network training status visualization: predicted rotating distributions of the agumentation network ({\bf Top}), loss distributions of pose network trained by adversarial ({\bf Middle}) and random ({\bf Bottom}) rotating augmentations. The augmentation network predicts rotating distributions matching the loss distributions of pose network, according to the first two rows. The loss distribution in the last row maintains a similar shape all the time due to the fixed Gaussian sampling distribution.}
\label{fig:distri}
\end{figure*}
\section{Experiments}\label{sec:experiments}
In this section, we first show the visualization of network training states to verify the motivation of doing adversarial dynamic augmentation. Then we quantitatively evaluate the effectiveness of different components in the method and further compare with state-of-the-art approaches. 

\subsection{Experiment Settings}
We use stacked hourglass \cite{newell2016stacked} as the pose network. The augmentation network takes the top-down part of an hourglass and only uses one cell module in each resolution block. To evaluate the generalization capability of the proposed adversarial augmentation, we tested two types of modules: {\it Residual module} \cite{he2016deep} and {\it Dense block} \cite{huang2016densely}.
The dense block provides direct connections among different layers, which helps the gradient flow in backpropagation.

{\bf Network design.}
We test both residual hourglass and dense hourglass in our component evaluation experiments. For residual hourglass, each residual module has a bottleneck structure of BN-ReLU-Conv(1x1)-BN-ReLU-Conv(3x3)-BN-ReLU-Conv(1x1). The input/output dimension of each bottleneck is 256. The two $1\times 1$ convolutions are used to halve and double the feature dimensions.

For dense hourglass, each module is a bottleneck structure of BN-ReLU-Conv(1x1)-BN-ReLU-Conv(3x3), with neck size 4, growth rate 32, and input dimension 128. The dimension increases by 32 after each dense layer. At the end of each dense block, we use BN-ReLU-Conv(1x1) to reduce the dimension to 128. We use the standard 8 stacked residual hourglasses \cite{newell2016stacked} as our baseline when compared with state-of-the-art methods.

{\bf Datasets.} We evaluate the proposed adversarial human pose estimation on two benchmark datasets: MPII Human Pose \cite{andriluka14cvpr} and  Leeds Sports Pose (LSP) \cite{johnson2010lsp}. MPII is collected from YouTube videos with a broad range of human activities. It has 25K images and 40K annotated persons (29K for training and 11K for testing). Following \cite{tompson2014joint}, we sample 3K samples from the training set for validation. Each person has 16 labeled joints. 

The LSP dataset contains images from many sports scenes. Its extended version has 11K training samples and 1K testing samples. Each person in LSP has 14 labeled joints. Since there are usually multiple people in one image, we crop around each person and resize it to $256\times 256$. Typically, random scaling (0.75-1.25), rotating (-/+30\textdegree) and flipping is used to augment the data.

{\bf Training.} We use PyTorch for the implementation. RMSProp \cite{tieleman2012rmsprop} is used to optimize the networks. The adversarial training contains three stages. We first train hourglass for a few epochs with a learning rate $2.5\times 10^{-4}$. Then we freeze the hourglass model and use it train the AHO and ASR networks with learning rate $2.5\times 10^{-4}$. Once they are pre-trained, we lower the learning rates of AHO and ASR networks to $5\times 10^{-5}$ and jointly train the three networks. The learning rate of the target network is decayed to $5\times 10^{-5}$ after the validation accuracy plateaus. In all experiments, the Percentage of Correct Keypoints (PCK) \cite{yang2011articulated} is used to measure the pose estimation accuracy.


\begin{table*}[th]
\centering
\caption{Comparison of random and adversarial data augmentation on the MPII validation set using PCKh@0.5.}
\vspace{-4pt}
\label{tab:soa_compare}.
\setlength\tabcolsep{3pt}
\begin{tabular}{l c c c c c c c c c c c c c c c c c}
\toprule
& \multicolumn{8}{c}{Residual hourglass (size: \textbf{38M})} & & \multicolumn{8}{c}{Dense hourglass (size: \textbf{18M})} \\
\cline{2-9} \cline{11-18} 
& Head & Sho. & Elb. & Wri. & Hip & Knee & Ank. & Mean & & Head & Sho. & Elb. & Wri. & Hip & Knee & Ank. & Mean \\
\cline{2-9} \cline{11-18}
Random Aug. &97.2 & 94.8 & 87.8 & 83.4 & 87.8 & 81.3 & 76.5 & 87.0 && 97.1 & 94.6 & 87.9 & 83.0 & 87.5 & 81.2 & 76.6 & 86.8\\
+ASR &97.3 & \textbf{95.2} & 88.2 & 84.2 & 88.2 & 81.8 & 77.3 & 87.5 && {\bf 97.2} & 95.0 & 88.3 & 83.5 & 87.7 & 81.8 & 77.4 & 87.3\\
+AHO &97.3 & 95.0 & 88.2 & 83.6 & 88.0 & 82.2 & 77.6 & 87.4 && 97.1 & 94.8 & 88.2 & 83.6 & 87.6 & 81.7 & 77.5 & 87.2\\
+ASR+AHO &\textbf{97.3} & 95.1 & \textbf{88.7} & \textbf{84.7} & \textbf{88.4} & \textbf{82.5} & \textbf{78.1} & \textbf{87.8} && \textbf{97.2} & \textbf{95.2} & \textbf{88.8} & \textbf{84.1} & \textbf{88.1} & \textbf{82.0} & \textbf{77.9} & \textbf{87.6}\\
\bottomrule
\end{tabular}\label{tb:componentEval}
\end{table*}

\begin{figure*}[ht]
\minipage{0.5\textwidth}
    \centering
  \includegraphics[width=0.95\linewidth]{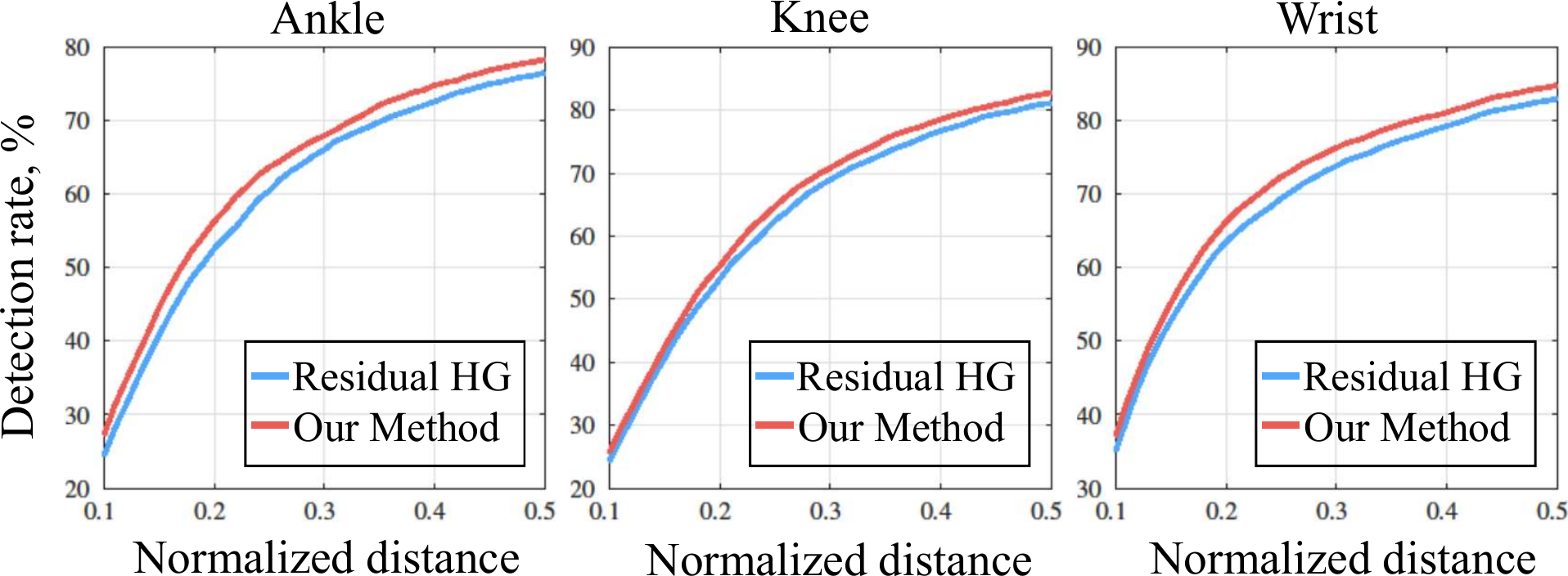}
\endminipage
\minipage{0.5\textwidth}
    \centering
  \includegraphics[width=0.95\linewidth]{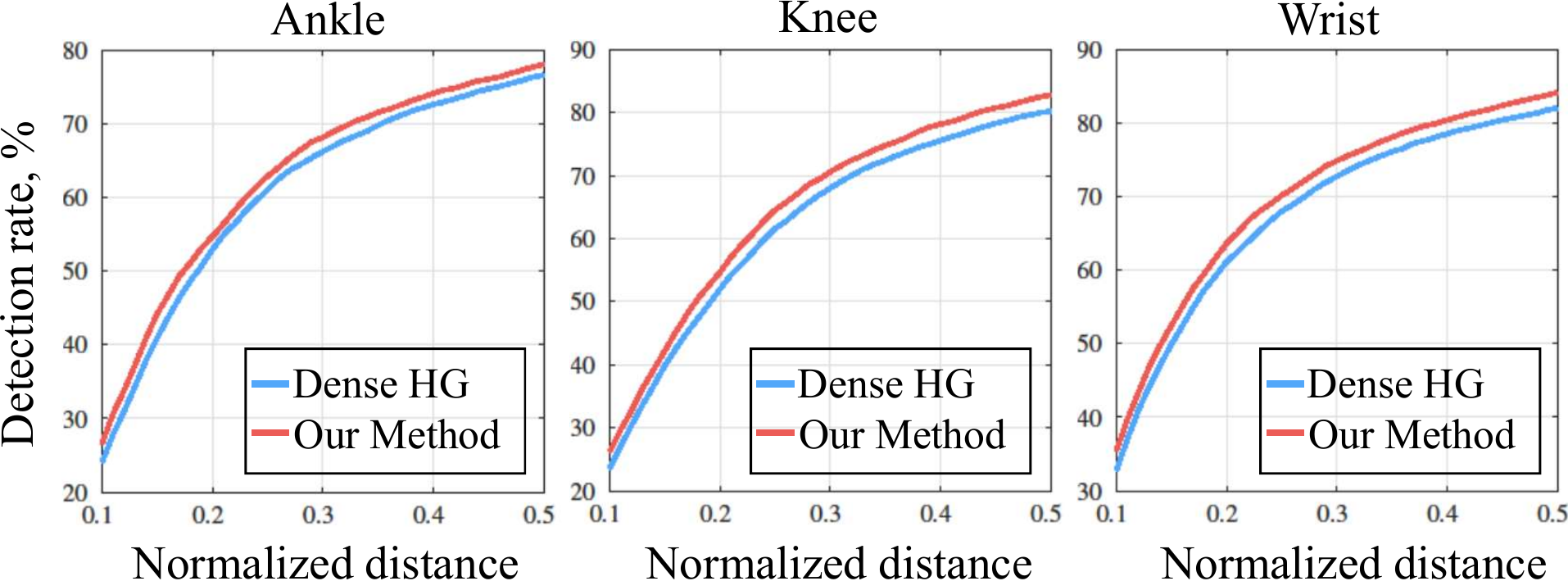}
\endminipage
\caption{Comparison of random and adversarial data augmentations on MPII validation set using PCKh@0.1-0.5. Consistent improvements on a range of normalized distances could be observed on both residual modules ({\bf left}) and dense blocks ({\bf right}).}
\label{fig:componentEval}
\end{figure*}

\subsection{Visualization of the Training Status}
In this experiment, we use a single residual hourglass. Each residual block contains 3 residual modules.
We are interested in knowing how the pose network handles human images with different data augmentations: rotating, scaling and occluding. Since our method treats these three variations in a similar way, we take rotating as an example. More specifically, we visualize the loss distribution of hourglass on images with different rotations.

{\bf Random data augmentation.} We train the pose network using random rotating sampled from a zero-centered Gaussian distribution as shown in the last row of Figure \ref{fig:distri}. We then test the trained pose network by applying the same rotating distribution on the testing data. We find that, at different training stages (training epochs), the target network loss always presents an inverted Gaussian-like distribution. 


{\bf Adversarial data augmentation.} In the beginning, the loss distribution of the pose network is similar to the case of random data augmentation. Since the pose network is pre-trained by the random data augmentation. However, the distribution becomes flatter as the training continues, which means the pose network could better handle the rotated images. The pose network learns from the adversarial data augmentation generated by the augmentation network.


{\bf Augmentation network training status.} The status can be visualized by applying the generated rotating augmentation. Comparing the first two rows in Figure \ref{fig:distri}, we can find that the generated rotating distribution is similar to the loss distribution of the pose network. This means that the augmentation network could track the training status of the target network and generate effective data augmentations.


\subsection{Component Evaluation}
We first verify the effectiveness of ASR and AHO in both residual and dense hourglasses. We use 3 residual bottlenecks in each block of residual hourglass. In dense hourglass, we use 6 densely connected bottlenecks in one dense block. Note that the size of dense hourglass model is less than half of the residual hourglass. In Table \ref{tb:componentEval}, we compare variants of adversarial data augmentation on PCKh@0.5. Figure \ref{fig:componentEval} shows the improvement of adversarial data augmentation compared with random data augmentation, on PCKh threshold from 0.1 to 0.5.

{\bf ASR only.} Table \ref{tb:componentEval} shows that ASR 
improves the accuracy of all the keypoints on both residual and dense hourglass, with average improvements of 0.5\% and 0.5\% respectively. This indicates that the generated adversarial scaling and rotating augmentations are effictive in training the pose network. 

{\bf AHO only.} Table \ref{tb:componentEval} shows that AHO 
improves accuracy on both residual and dense hourglass, with average improvements of 0.4\% and 0.4\% respectively. Similarly, the pose network can also learn improved inference from the adversarial occluding generated by the augmentation network.

{\bf ASR and AHO.} Applying both ASR and AHO can further improve the accuracy by 0.4\%, compared with applying either of them. Figure \ref{fig:componentEval} shows that ASR and AHO can significantly improve the localization accuracy especially for joints that are usually more difficult to localize, such as ankle, knee and wrist. 

{\bf Dense hourglass vs Residual hourglass.} Table \ref{tb:componentEval} also shows that the dense hourglass has comparable performance in terms of pose estimation accuracy, but much more parameter efficient than the residual design (18M {\it vs.} 38M). The dense design facilitates the gradient flow through the direct connections among different feature blocks, which uses fewer parameters without sacrificing the estimation accuracy.

\subsection{Comparing with State-of-the-art Methods}
\begin{figure*}[hbt]
\centering
  \includegraphics[width=\linewidth]{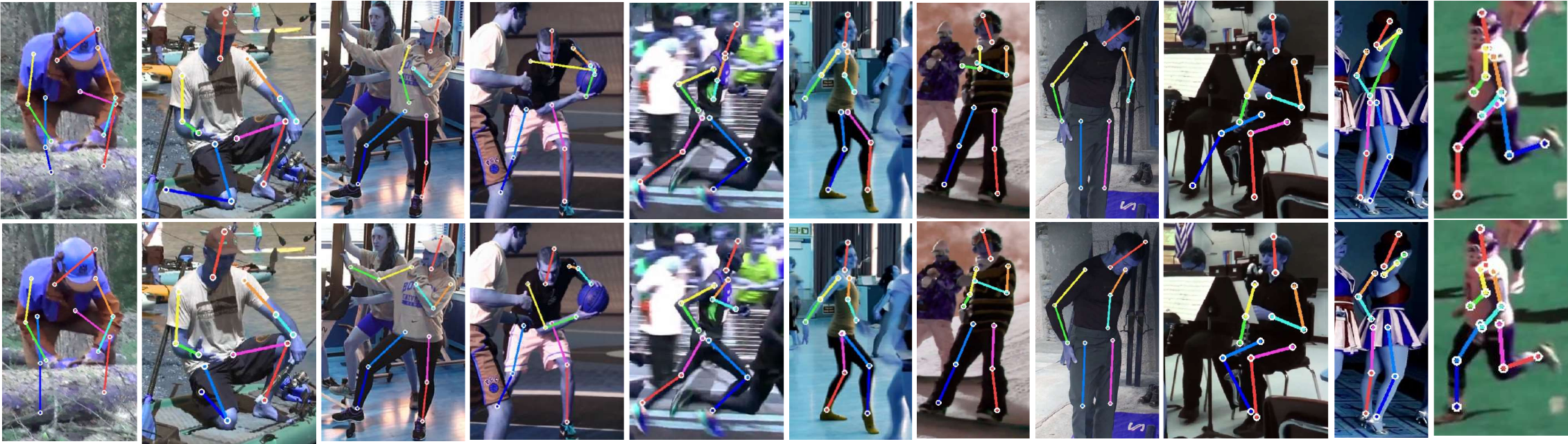}
\caption{Comparisons of the same Stacked HG network trained using random data augmentation ({\bf top}) and adversarial data augmentation ({\bf bottom}). Note the improvement on challenging joints ({\it e.g.} ankle, elbow, wrist), and left-right confusion.}
\label{fig:qualitatve}
\end{figure*}

{\bf Quantitative comparison.} 
To compare with state-of-the-art methods, we apply the proposed adversarial data augmentation to train the hourglasses network (totally 8 stacked) \cite{newell2016stacked}. 
The bridge features generated by the first hourglass network in the stack are input into the adversarial network.
The same hierarchical occluding masks are applied to every hourglass network in the stack.
Table \ref{tb:MPII} compares PCKh@0.5 accuracy of different methods on MPII dataset.
The proposed adversarial data augmentation can improve the baseline \cite{newell2016stacked} by 0.6\%, which achieves state-of-the-art performance. 
Table \ref{tb:LSP} compares PCK@0.2 accuracy of different methods on LSP dataset. Again, our method can improve the baseline \cite{newell2016stacked} by 1.5\%, which significantly outperforms state-of-the-art methods.

{\bf Qualitative Comparison.} Figure \ref{fig:qualitatve} shows qualitative comparisons. We compare the random and adversarial data augmentation. We can observer the improvement resulted from the adversarial data augmentation. Interestingly, the pose network could handle the left-right confusions after the adversarial training.

\begin{table}[htb]
\begin{center}
\caption{PCKh@0.5 on the MPII test set. Our adversarial data augmentation improves baseline stacked HGs(8) \cite{newell2016stacked}.}
\label{tb:MPII}
\small
\setlength\tabcolsep{1.5pt}
\begin{tabular}{@{}lcccccccc@{}}
\toprule
Method & Head & Sho. & Elb. & Wri. & Hip & Knee & Ank. & Mean\\
\hline
Pishchulin \textit{et al.}\cite{pishchulin2013strong} & 74.3 & 49.0 & 40.8 & 34.1 & 36.5 & 34.4 & 35.2 & 44.1\\
Tompson \textit{et al.}\cite{tompson2014joint} & 95.8 & 90.3 & 80.5 & 74.3 & 77.6 & 69.7 & 62.8 & 79.6\\
Carreira \textit{et al.}\cite{carreira2016human} & 95.7 & 91.7 & 81.7 & 72.4 & 82.8 & 73.2 & 66.4 & 81.3\\
Tompson \textit{et al.}\cite{tompson2015efficient}& 96.1 & 91.9 & 83.9 & 77.8 & 80.9 & 72.3 & 64.8 & 82.0\\
Hu \textit{et al.}\cite{hu2016bottom}& 95.0 & 91.6 & 83.0 & 76.6 & 81.9 & 74.5 & 69.5 & 82.4\\
Pishchulin \textit{et al.}\cite{pishchulin2016deepcut}&94.1 & 90.2 & 83.4 & 77.3 & 82.6 & 75.7 & 68.6 & 82.4\\
Lifshitz \textit{et al.}\cite{lifshitz2016human} & 97.8 & 93.3 & 85.7 & 80.4 & 85.3 & 76.6 & 70.2 & 85.0\\
Gkioxary \textit{et al.}\cite{gkioxari2016chained} & 96.2 & 93.1 & 86.7 & 82.1 & 85.2 & 81.4 & 74.1 & 86.1\\
Rafi \textit{et al.}\cite{rafi2016efficient} & 97.2 & 93.9 & 86.4 & 81.3 & 86.8 & 80.6 & 73.4 & 86.3\\
Belagiannis \textit{et al.}\cite{belagiannis2017recurrent}&97.7 & 95.0 & 88.2 & 83.0 & 87.9 & 82.6 & 78.4 & 88.1\\
Insafutdinov \textit{et al.}\cite{insafutdinov2016deepercut}&96.8 & 95.2 & 89.3 & 84.4 & 88.4 & 83.4 & 78.0 & 88.5\\
Wei \textit{et al.}\cite{wei2016convolutional} & 97.8 & 95.0 & 88.7 & 84.0 & 88.4 & 82.8 & 79.4 & 88.5\\
Bulat \textit{et al.}\cite{bulat2016human} & 97.9 & 95.1 & 89.9 & 85.3 & 89.4 & 85.7 & 81.7 & 89.7\\
Chu \textit{et al.}\cite{chu2017multi} & {\bf 98.5} & 96.3 & 91.9 & 88.1 & 90.6 & {\bf 88.0} & {\bf 85.0} & {\bf 91.5}\\
\hline
Stacked HGs(8) \cite{newell2016stacked} & 98.2 & 96.3 & 91.2 & 87.1 & 90.1 & 87.4 & 83.6 & 90.9\\
Ours: +ASR+AHO & 98.1  & {\bf 96.6}  & {\bf 92.5}  & {\bf 88.4}  & {\bf 90.7}  & 87.7 & 83.5 & {\bf 91.5}\\
\bottomrule
\end{tabular}
\end{center}
\vspace{-10pt}
\end{table}

\begin{table}[htb]
\begin{center}
\caption{PCK@0.2 on the LSP dataset. Clear improvements are observed over the baseline stacked HGs(8) \cite{newell2016stacked}.}
\label{tb:LSP}
\small
\setlength\tabcolsep{1.5pt}
\begin{tabular}{@{}lcccccccc@{}}
\toprule
Method & Head & Sho. & Elb. & Wri. & Hip & Knee & Ank. & Mean\\
\hline
Belagiannis \textit{et al.}\cite{belagiannis2017recurrent} & 95.2 & 89.0 & 81.5 & 77.0 & 83.7 & 87.0 & 82.8 & 85.2\\
Lifshitz \textit{et al.}\cite{lifshitz2016human} & 96.8 & 89.0 & 82.7 & 79.1 & 90.9 & 86.0 & 82.5 & 86.7\\
Pishchulin \textit{et al.}\cite{pishchulin2016deepcut} &  97.0 & 91.0 & 83.8 & 78.1 & 91.0 & 86.7 & 82.0 & 87.1\\
Insafutdinov \textit{et al.}\cite{insafutdinov2016deepercut}& 97.4 & 92.7 & 87.5 & 84.4 & 91.5 & 89.9 & 87.2 & 90.1\\
Wei \textit{et al.}\cite{wei2016convolutional}& 97.8 & 92.5 & 87.0 & 83.9 & 91.5 & 90.8 & 89.9 & 90.5\\
Bulat \textit{et al.}\cite{bulat2016human}& 97.2 & 92.1 & 88.1 & 85.2 & 92.2 & 91.4 & 88.7 & 90.7\\
Chu \textit{et al.}\cite{chu2017multi}& 98.1 & 93.7 & 89.3 & 86.9 &  93.4 & 94.0 & 92.5 & 92.6\\
\hline
Stacked HGs(8) \cite{newell2016stacked} & 98.2 & 94.0 & 91.2 & 87.2 & 93.5 & 94.5 & 92.6 & 93.0\\
\textbf{Ours: ASR+AHO} & \textbf{98.6} & \textbf{95.3} & \textbf{92.8} & \textbf{90.0} & \textbf{94.8} & \textbf{95.3} & \textbf{94.5} & \textbf{94.5}\\
\bottomrule
\end{tabular}
\end{center}
\vspace{-10pt}
\end{table}



\vspace{-0.1in}
\section{Conclusion}
In this paper, we proposed a new method to jointly optimize data augmentation and network training. An augmentation network is designed to generate adversarial data augmentations in order to improve the training of a target network. Improved performance has been observed by applying our method on human pose estimation. In the future, we plan to further improve our method for more general applications in visual and language understanding.

\section{Acknowledgment}
This work is partly supported by the Air Force Office of Scientific Research (AFOSR) under the Dynamic Data-Driven Application Systems program. We also thank to NSF Computer and Information Science and Engineering (CISE) for the support of our research.

{\small
\bibliographystyle{ieee}
\bibliography{egbib}

\begin{thebibliography}{10}\itemsep=-1pt

\bibitem{andriluka14cvpr}
M.~Andriluka, L.~Pishchulin, P.~Gehler, and B.~Schiele.
\newblock {2D} human pose estimation: New benchmark and state of the art
  analysis.
\newblock In {\em CVPR}, 2014.

\bibitem{belagiannis2017recurrent}
V.~Belagiannis and A.~Zisserman.
\newblock Recurrent human pose estimation.
\newblock In {\em FG}, 2017.

\bibitem{bulat2016human}
A.~Bulat and G.~Tzimiropoulos.
\newblock Human pose estimation via convolutional part heatmap regression.
\newblock In {\em ECCV}, 2016.

\bibitem{carreira2016human}
J.~Carreira, P.~Agrawal, K.~Fragkiadaki, and J.~Malik.
\newblock Human pose estimation with iterative error feedback.
\newblock In {\em CVPR}, 2016.

\bibitem{chen2014articulated}
X.~Chen and A.~L. Yuille.
\newblock Articulated pose estimation by a graphical model with image dependent
  pairwise relations.
\newblock In {\em NIPS}, 2014.

\bibitem{yu2017adversarial}
Y.~Chen, C.~Shen, X.-S. Wei, L.~Liu, and J.~Yang.
\newblock Adversarial posenet: A structure-aware convolutional network for
  human pose estimation.
\newblock In {\em ICCV}, 2017.

\bibitem{chou2017self}
C.-J. Chou, J.-T. Chien, and H.-T. Chen.
\newblock Self adversarial training for human pose estimation.
\newblock {\em arXiv}, 2017.

\bibitem{chu2017multi}
X.~Chu, W.~Yang, W.~Ouyang, C.~Ma, A.~L. Yuille, and X.~Wang.
\newblock Multi-context attention for human pose estimation.
\newblock {\em arXiv}, 2017.

\bibitem{Mohamed17}
M.~Elhoseiny, Y.~Zhu, H.~Zhang, and A.~Elgammal.
\newblock Link the head to the “beak”: Zero shot learning from noisy text
  description at part precision.
\newblock In {\em CVPR}, 2017.

\bibitem{girshick2014rich}
R.~Girshick, J.~Donahue, T.~Darrell, and J.~Malik.
\newblock Rich feature hierarchies for accurate object detection and semantic
  segmentation.
\newblock In {\em CVPR}, 2014.

\bibitem{gkioxari2016chained}
G.~Gkioxari, A.~Toshev, and N.~Jaitly.
\newblock Chained predictions using convolutional neural networks.
\newblock In {\em ECCV}, 2016.

\bibitem{goodfellow2014gan}
I.~J. Goodfellow, J.~Pouget{-}Abadie, M.~Mirza, B.~Xu, D.~Warde{-}Farley,
  S.~Ozair, A.~C. Courville, and Y.~Bengio.
\newblock Generative adversarial nets.
\newblock In {\em {NIPS}}, 2014.

\bibitem{he2017maskrcnn}
K.~He, G.~Gkioxari, P.~Dollár, and R.~Girshick.
\newblock {Mask R-CNN}.
\newblock In {\em ICCV}, 2017.

\bibitem{he2016deep}
K.~He, X.~Zhang, S.~Ren, and J.~Sun.
\newblock Deep residual learning for image recognition.
\newblock In {\em CVPR}, 2016.

\bibitem{hu2016bottom}
P.~Hu and D.~Ramanan.
\newblock Bottom-up and top-down reasoning with hierarchical rectified
  gaussians.
\newblock In {\em CVPR}, 2016.

\bibitem{huang2016densely}
G.~Huang, Z.~Liu, K.~Q. Weinberger, and L.~van~der Maaten.
\newblock Densely connected convolutional networks.
\newblock {\em arXiv}, 2016.

\bibitem{insafutdinov2016deepercut}
E.~Insafutdinov, L.~Pishchulin, B.~Andres, M.~Andriluka, and B.~Schiele.
\newblock Deepercut: A deeper, stronger, and faster multi-person pose
  estimation model.
\newblock In {\em ECCV}, 2016.

\bibitem{jaderberg2015spatial}
M.~Jaderberg, K.~Simonyan, A.~Zisserman, et~al.
\newblock Spatial transformer networks.
\newblock In {\em NIPS}, 2015.

\bibitem{johnson2010lsp}
S.~Johnson and M.~Everingham.
\newblock Clustered pose and nonlinear appearance models for human pose
  estimation.
\newblock In {\em BMVC}, 2010.

\bibitem{alex2012alexnet}
A.~Krizhevsky, I.~Sutskever, and G.~E. Hinton.
\newblock Imagenet classification with deep convolutional neural networks.
\newblock In {\em {NIPS}}, 2012.

\bibitem{lecun1998gradient}
Y.~LeCun, L.~Bottou, Y.~Bengio, and P.~Haffner.
\newblock Gradient-based learning applied to document recognition.
\newblock {\em IEEE}, 1998.

\bibitem{lifshitz2016human}
I.~Lifshitz, E.~Fetaya, and S.~Ullman.
\newblock Human pose estimation using deep consensus voting.
\newblock In {\em ECCV}, 2016.

\bibitem{long2015fully}
J.~Long, E.~Shelhamer, and T.~Darrell.
\newblock Fully convolutional networks for semantic segmentation.
\newblock In {\em CVPR}, 2015.

\bibitem{newell2016stacked}
A.~Newell, K.~Yang, and J.~Deng.
\newblock Stacked hourglass networks for human pose estimation.
\newblock In {\em ECCV}, 2016.

\bibitem{peng2016recurrent}
X.~Peng, R.~S. Feris, X.~Wang, and D.~N. Metaxas.
\newblock A recurrent encoder-decoder network for sequential face alignment.
\newblock In {\em ECCV}, 2016.

\bibitem{peng2015circle}
X.~Peng, J.~Huang, Q.~Hu, S.~Zhang, A.~Elgammal, and D.~Metaxas.
\newblock From circle to 3-sphere: Head pose estimation by instance
  parameterization.
\newblock {\em CVIU}, 2015.

\bibitem{pishchulin2013strong}
L.~Pishchulin, M.~Andriluka, P.~Gehler, and B.~Schiele.
\newblock Strong appearance and expressive spatial models for human pose
  estimation.
\newblock In {\em ICCV}, 2013.

\bibitem{pishchulin2016deepcut}
L.~Pishchulin, E.~Insafutdinov, S.~Tang, B.~Andres, M.~Andriluka, P.~V. Gehler,
  and B.~Schiele.
\newblock Deepcut: Joint subset partition and labeling for multi person pose
  estimation.
\newblock In {\em CVPR}, 2016.

\bibitem{rafi2016efficient}
U.~Rafi, B.~Leibe, J.~Gall, and I.~Kostrikov.
\newblock An efficient convolutional network for human pose estimation.
\newblock In {\em BMVC}, 2016.

\bibitem{reed2016generative}
S.~Reed, Z.~Akata, X.~Yan, L.~Logeswaran, B.~Schiele, and H.~Lee.
\newblock Generative adversarial text to image synthesis.
\newblock {\em arXiv}, 2016.

\bibitem{ronneberger2015unet}
O.~Ronneberger, P.~Fischer, and T.~Brox.
\newblock U-net: Convolutional networks for biomedical image segmentation.
\newblock In {\em {MICCAI}}, Lecture Notes in Computer Science, 2015.

\bibitem{shrivastava2016training}
A.~Shrivastava, A.~Gupta, and R.~Girshick.
\newblock Training region-based object detectors with online hard example
  mining.
\newblock In {\em CVPR}, 2016.

\bibitem{tang2015face}
Z.~Tang, Y.~Zhang, Z.~Li, and H.~Lu.
\newblock Face clustering in videos with proportion prior.
\newblock In {\em IJCAI}, 2015.

\bibitem{tieleman2012rmsprop}
T.~Tieleman and G.~Hinton.
\newblock Lecture 6.5-rmsprop: Divide the gradient by a running average of its
  recent magnitude.
\newblock In {\em NNML}, 2012.

\bibitem{tompson2015efficient}
J.~Tompson, R.~Goroshin, A.~Jain, Y.~LeCun, and C.~Bregler.
\newblock Efficient object localization using convolutional networks.
\newblock In {\em CVPR}, 2015.

\bibitem{tompson2014joint}
J.~J. Tompson, A.~Jain, Y.~LeCun, and C.~Bregler.
\newblock Joint training of a convolutional network and a graphical model for
  human pose estimation.
\newblock In {\em NIPS}, 2014.

\bibitem{toshev2014deeppose}
A.~Toshev and C.~Szegedy.
\newblock Deeppose: Human pose estimation via deep neural networks.
\newblock In {\em CVPR}, 2014.

\bibitem{uijlings2013selective}
J.~R. Uijlings, K.~E. Van De~Sande, T.~Gevers, and A.~W. Smeulders.
\newblock Selective search for object recognition.
\newblock {\em IJCV}, 2013.

\bibitem{wang2017fast}
X.~Wang, A.~Shrivastava, and A.~Gupta.
\newblock A-fast-rcnn: Hard positive generation via adversary for object
  detection.
\newblock {\em arXiv}, 2017.

\bibitem{wei2016convolutional}
S.-E. Wei, V.~Ramakrishna, T.~Kanade, and Y.~Sheikh.
\newblock Convolutional pose machines.
\newblock In {\em CVPR}, 2016.

\bibitem{wu2016towards}
L.~Wu, K.~J. Wu, A.~Sim, M.~Churchill, J.~Y. Choi, A.~Stathopoulos, C.-S.
  Chang, and S.~Klasky.
\newblock Towards real-time detection and tracking of spatio-temporal features:
  Blob-filaments in fusion plasma.
\newblock {\em TBD}, 2016.

\bibitem{yang2017learning}
W.~Yang, S.~Li, W.~Ouyang, H.~Li, and X.~Wang.
\newblock Learning feature pyramids for human pose estimation.
\newblock {\em arXiv}, 2017.

\bibitem{yang2011articulated}
Y.~Yang and D.~Ramanan.
\newblock Articulated pose estimation with flexible mixtures-of-parts.
\newblock In {\em CVPR}, 2011.

\bibitem{yu2017semantic}
A.~Yu and K.~Grauman.
\newblock Semantic jitter: Dense supervision for visual comparisons via
  synthetic images.
\newblock In {\em ICCV}, 2017.

\bibitem{zhang2017image}
H.~Zhang, V.~Sindagi, and V.~M. Patel.
\newblock Image de-raining using a conditional generative adversarial network.
\newblock {\em arXiv}, 2017.

\bibitem{zhu2014capturing}
X.~Zhu, D.~Anguelov, and D.~Ramanan.
\newblock Capturing long-tail distributions of object subcategories.
\newblock In {\em CVPR}, 2014.

\bibitem{yizhe_zsl_2018}
Y.~Zhu, M.~Elhoseiny, B.~Liu, X.~Peng, and A.~Elgammal.
\newblock A generative adversarial approach for zero-shot learning from noisy
  texts.
\newblock In {\em CVPR}, 2018.

\end{thebibliography}
}

\end{document}